\documentclass[runningheads]{llncs}
\usepackage[T1]{fontenc}
\usepackage{graphicx,verbatim}

\usepackage{multirow}
\usepackage[table,xcdraw]{xcolor}
\usepackage{makecell}
\usepackage{array}
\usepackage{caption}
\usepackage{colortbl}
\usepackage{ragged2e}
\usepackage{rotating}

\usepackage{ulem}

\usepackage[utf8]{inputenc}

\usepackage{amsmath,amssymb,graphicx}
\usepackage{mathtools,bm}

\usepackage{orcidlink}
\usepackage{tikz}

\usepackage{amsmath}
\DeclareMathOperator*{\argmax}{arg\,max}

\DeclareMathOperator*{\argsort}{arg\,sort}

\usepackage[dvipsnames]{xcolor}
\usepackage{xcolor}
\definecolor{myGold}{RGB}{240, 221, 168}

\newcommand{\crule}[3][black]{\textcolor{#1}{\rule{#2}{#3}}}

\begin{document}
\title{RaLMPH: Reliability-aware Learning for Multi-Pathologist Harmonization in\\ Whole-Slide Image Classification}
\titlerunning{RaLMPH}
%

\author{Sungrae Hong\inst{1}\orcidlink{0009-0009-9653-6662} \and
Jiwon Jeong\inst{1}\orcidlink{0009-0005-7535-3104} \and
Soeun Cheon\inst{1}\orcidlink{0009-0001-9384-8742} \and
Donghee Han\inst{1}\orcidlink{0000-0002-4560-6810} \and \\
Sol Lee\inst{1}\orcidlink{0009-0004-6520-2137} \and
Jisu Shin\inst{1}\orcidlink{0009-0008-8256-4788} \and
Kyungeun Kim\inst{2}\orcidlink{0000-0001-7938-4673} \and
Mun Yong Yi\thanks{Corresponding Author}\inst{1}\orcidlink{0000-0003-1784-8983}}


\authorrunning{Hong et al.}

\institute{Korea Advanced Institute of Science and Technology, Daejeon, South Korea\\
\email{\{sr5043, zzioni, eintausend, handonghee,\\leeson4553, jisu3389, munyi\}@kaist.ac.kr} \and
Seegene Medical Foundation, Seoul, South Korea\\
\email{kekim@mf.seegene.com}}

  
\maketitle              
\begin{abstract}
%
Multiple Instance Learning (MIL) is a standard paradigm for Whole-Slide Image (WSI) analysis and has achieved strong results in computational pathology. However, most MIL pipelines assume a single “gold” label per slide, which conflicts with clinical practice where substantial inter-pathologist variability is common. Existing multi-annotator learning and label-refinement methods typically estimate global annotator reliability or rely on single-instance assumptions, making them poorly suited to MIL and to localized diagnostic contexts where experts disagree. We propose RaLMPH (Reliability-aware Learning for Multi-Pathologist Harmonization), a MIL-based label reconciliation framework for WSIs annotated by multiple pathologists. RaLMPH introduces a reliability field that jointly models (i) local neighborhood structure in WSI feature space and (ii) expert uncertainty (entropy), enabling per-sample identification of trustworthy reference neighborhoods. Leveraging this field, RaLMPH performs sample-wise local annotator ranking to select reliable opinions per slide and applies an adaptive gating mechanism to fuse labels conditioned on local reliability. Experiments on a clinical WSI dataset with labels from six pathologists, as well as controlled simulated benchmarks, show that RaLMPH consistently outperforms existing approaches. Further analyses clarify how our reliability-aware mechanism improves label reconciliation and downstream MIL performance.
%
\end{abstract}

\section{Introduction}
In response to the surge in demand for pathology specimen diagnosis following the COVID-19 pandemic~\cite{bychkov2023constant}, the deep learning (DL) community has heavily leveraged Multiple Instance Learning (MIL)~\cite{zhang2025patches}. MIL utilizes weakly supervised Whole-Slide Image (WSI) level labels, significantly alleviating the annotation burden for medical experts, i.e., pathologists~\cite{gadermayr2024multiple}.

Conventional MIL studies presuppose idealized benchmarks where a single, unambiguous label is assigned to each WSI~\cite{sudharshan2019multiple}. However, achieving consensus on the WSI label among experts in a real-world clinic is challenging due to diagnostic ambiguity and differences in their experience and training~\cite{kweldam2016gleason,montgomery2001reproducibility}, which represents a spectrum of expert judgment arising from disease complexity rather than mere noise. Therefore, the application of existing MIL methodologies requires a label integration process via collaborative consultation and consensus diagnosis~\cite{brancati2022bracs}. Nonetheless, it poses a practically infeasible constraint, as it would require a limited cohort of experts to engage in time-intensive deliberations to reach consensus on the vast volume of samples required for MIL training.

Various attempts have been made to utilize multiple labels assigned to a single sample to train DL models~\cite{uma2021learning}. Majority voting (MV) and soft labeling (SL) are the most intuitive methods~\cite{sheng2008get}. To overcome the inherent limitation that these are not robust to unreliable annotators, hidden state-based methods were proposed~\cite{chu2021learning,whitehill2009whose}. Advances in DL, which involve large-scale parameter optimization, have enabled approaches that directly learn robust label distributions~\cite{guan2018said,rodrigues2018deep}. Recently, it has focused on the representation of input is correlated with annotator disagreement, leading to new approach that leverage features for label refinement~\cite{jiang2021learning,li2023neighborhood}.

Nevertheless, previous approaches fall short of addressing the unique challenges of training MIL in a real-world clinic. For analysis, a gigapixel WSI is partitioned into thousands of patches and aggregated into an instance bag~\cite{zhang2025patches}. Given that existing approaches are grounded in the assumption of single-instance data points~\cite{jiang2021learning,li2023neighborhood,zhang2024kfnn}, their application to multi-instance bags of WSI remains fundamentally restricted. 
Furthermore, unlike previous belief that specific annotators are globally (un)reliable~\cite{chu2021learning,whitehill2009whose,zhou2012learning}, the nature of clinical WSIs, where labels are provided by pathology experts, suggests that it should not be viewed as global inferiority~\cite{elmore2015diagnostic}. Instead, it is interpreted as a reflection of specialized expertise that fluctuates across localized diagnostic points, determining which expert holds the highest clinical authority for a given sample~\cite{elmore2017pathologists,van1991understanding}.

\begin{figure}[t!]
\centering
\includegraphics[width=1\linewidth]{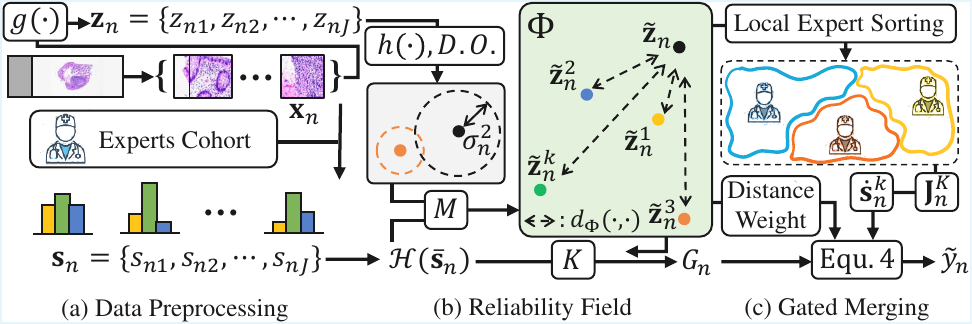}
\caption{
Overview of the RaLMPH. (a) An expert cohort provides multiple labels for each sample. (b) The reliability field integrates the variance of the instance bag with the multiple label entropy. (c) Labels are reconciled, weighted by locally reliable experts.
}
\end{figure}\label{fig:overview}

To tackle these challenges, which have remained unexplored venue, we propose \textbf{R}eliability-\textbf{a}ware \textbf{L}earning for \textbf{M}ulti-\textbf{P}athologist \textbf{H}armonization (RaLMP\\H), the label reconciliation for WSIs annotated by multiple experts. We introduce a reliability field that merges input and label uncertainties, which maps reliable neighbors by jointly modeling WSI ambiguity and expert non-consensus.
Recognizing that pathological labels are expert-driven, we move beyond global reliability to selecting sample-wise reliable experts. The proposed gating determines the degree of label merging based on the local entropy of defined neighbors, ensuring that expert opinions are aggregated only where local consensus is observed. Experiments conducted on clinical WSI labeled by six pathology experts, along with simulated benchmarks, demonstrate that the proposed method is both a practical and a generalized approach, while RaLMPH consistently outperforms competitive methods. Various in-depth analyses provide a comprehensive understanding of the RaLMPH.
\section{Method}
Let $\mathbf{D}=\{(\mathbf{x}_n,\mathbf{s}_n)\}_{n=1}^N$ denotes a WSI dataset, where $N$ is the number of samples, and each instance bag $\mathbf{x}_n=\{x_{n1},\cdots,x_{n|\mathbf{x}_n|}\}$ is annotated by $J$ experts corhort, i.e., $\mathbf{s}_n=\{s_{n1},\cdots,s_{nJ}\}$. Expert $j$ assigns a soft label $s_{nj}\in\mathbb{R}^C$, where $C$ is the number of classes, only to the $\mathbf{x}_n$, not to individual instances $x_{n\mathbf{\cdot}}$. A pathology foundation model~\cite{lu2024avisionlanguage} extracts the feature $g(x_{n\cdot})=z_{n\cdot}\in\mathbb{R}^D$, which organizes a feature bag $\mathbf{z}_n=\{z_{n1},\cdots,z_{n|\mathbf{x}_n|}\}$ and $D$ is embedding dimension (see Fig.~\ref{fig:overview}(a)). The proposed method aims to reconcile WSI labels from multiple experts, thereby enabling the training of a general MIL model $f_\theta(\mathbf{z}_n)$.

\subsection{Reliability Field}
Recent works have highlighted that label non-consensus stems not only from subjective annotator differences but also from the inherent properties of the samples themselves~\cite{jiang2021learning,li2023neighborhood,zhang2024kfnn}. However, since a WSI feature bag $\mathbf{z}_n$ consists of thousands of instances, conventional methods designed for single data points cannot be directly applied. Therefore, we leverage a WSI-level feature extractor~\cite{jaume2024multistain}, i.e., $h(\mathbf{z}_n)=\tilde{\mathbf{z}}_n\in\mathbb{R}^D$. While we utilize clustering for comprehensive sample analysis, the $\tilde{\mathbf{z}}_n$ often provides an insufficiently descriptive representation~\cite{zhang2024disease}. Therefore, we define the reliability field $\Phi$ as Fig.~\ref{fig:overview}(b) and Eq.~\ref{eq:distance}:
\begin{equation}
    \begin{aligned}\label{eq:distance}
    d_{\Phi}(\mathbf{z}_n, \mathbf{z}_m) = \sqrt{(\tilde{\mathbf{z}}_n - \tilde{\mathbf{z}}_m)^\top \mathbf{M}_{nm} (\tilde{\mathbf{z}}_n - \tilde{\mathbf{z}}_m)}
    \mspace{130mu}
    \\
    \text{, where } \mathbf{M}_{nm} = \text{diag}\left( \exp \left(  {\mathcal{H}(\bar{\mathbf{s}}_n) \cdot \bm{\sigma}^2_n + \mathcal{H}(\bar{\mathbf{s}}_m) \cdot \bm{\sigma}^2_m} \right) \right)\in\mathbb{R}^{D\times{D}}.
    \mspace{25mu}
    \end{aligned}
\end{equation}
The reliability field $\Phi$ integrates the expert opinion space and the variance space of the feature bag into the distance $d_\Phi(\cdot,\cdot)$, pushing ambiguous samples further apart. $\mathbf{M}_{nm}$ is a weight matrix that determines the reliability of information, assigning higher values as label entropy $\mathcal{H}(\bar{\mathbf{s}}_\cdot)$ and sample atypicality increase. In practice, $\mathbf{\bar{s}}_n=(1/J)\times\sum_{j=1}^J{s_{nj}}$, and variance $\bm\sigma_n^2\in\mathbb{R}^D$ is measured by applying feature dropout~\cite{park2025uncertainty} multiple times to $\mathbf{z}_n$.

\subsection{Gated Label Reconciling via Locally Reliable Expert Sorting}
The interaction between sample and label, i.e., $\bm\sigma_n^2$ and cohort entropy $\mathcal{H}(\mathbf{\bar{s}}_n)$, roughly manifests in four scenarios:  low (high) $\bm\sigma_n^2$ $\times$ low (high) $\mathcal{H}(\mathbf{\bar{s}}_n)$. Cases with only low $\mathcal{H}(\mathbf{\bar{s}}_n)$ can be considered highly reliable $\mathbf{\bar{s}}_n$, as experienced medical experts provide the labels with high consensus. Conversely, high $\mathcal{H}(\mathbf{\bar{s}}_n)$ cases necessitate examining the relationship between the $\mathbf{z}_n$ and its corresponding neighbors. To formalize this conditional logic, we employ a gating function as:
\begin{equation}
    G_n = \text{sigmoid} \left(\exp(1/\mathcal{H}(\bar{\mathbf{s}}_n))+1/\exp(1/\bar{\mathcal{H}} (\bar{\mathbf{s}}_{K}))-1 \right)\in(0.5,1),
\end{equation}\label{eq:gating}
where $\bar{\mathcal{H}} (\bar{\mathbf{s}}_{K})=(1/K)\times\sum_{k{\in}K}\mathcal{H}(\bar{\mathbf{s}}_k)$ is the average entropy of the labels of the neighbors $K$ of $\mathbf{z}_n$. $G_n$ is close to 1 in the cases of low $\mathcal{H}(\mathbf{\bar{s}}_n)$, i.e., high consensus, while low $\bar{\mathcal{H}} (\bar{\mathbf{s}}_{K})$ determines $G_n$ close to $0.5$ when $\mathcal{H}(\mathbf{\bar{s}}_n)$ is high. In cases where $\bar{\mathcal{H}} (\bar{\mathbf{s}}_{K})$ also lacks in consensus, the $G_n$ prioritizes the initial label of $\mathbf{x}_n$ to drive it toward $\text{sigmoid}(1) \approx 0.73$.

Unlike previous approaches, which involve non-experts~\cite{guan2018said,li2023neighborhood}, all clinical specialist annotators can be considered locally reliable candidates~\cite{elmore2015diagnostic}. We therefore propose a local expert sorting to formulate the expert reliability for each $\mathbf{z}_n$:
\begin{equation}
    \mathbf{J}^{K}_n =\argsort_{j\in{J}} \sum_{k \in K}\text{cosine-sim} ({s}_{kj}, \bar{\mathbf{s}}_k)\in\mathbb{R}^{J}.
\end{equation}\label{eq:argsort}
$\mathbf{J}^{K}_n$ is a list that contains the indices of the reliable local experts' ranking.

The reconciled label $\tilde{y}_n$ within the $\Phi$, where Fig.~\ref{fig:overview}(c) deficts, is defined as:
\begin{equation}
   \begin{aligned}\label{eq:label}
    \tilde{y}_n = G_n \times \bar{\mathbf{s}}_n + (1-G_n) \times \left( \sum_{k \in K^*} 
    W_k \cdot
    \dot{\mathbf{s}}^k_n \right)
    \in \mathbb{R}^C
    \\
    \text{, where }
    \begin{cases*}
        W_k=\frac{{\exp}\left( -d_{\Phi}(\mathbf{z}_n,\mathbf{z}_n^k)/\tau\right)}{\sum_{k'\in K^*}{\exp}\left( -d_{\Phi}(\mathbf{z}_n,\mathbf{z}_n^{k'})/\tau\right)}
        \\
        \dot{\mathbf{s}}^k_n=\sum_{j=1}^{J}\left(\frac{\exp(J-j+1)}{\sum_{j'=1}^{J}\exp(j')}s_{k\mathbf{J}_n^K[j]}\right).
    \end{cases*}
    \mspace{30mu}
    \end{aligned}
\end{equation}
When the $\bar{\mathbf{s}}_n$ is ambiguous, it performs a weighted reconciling of the $K^*$ nearest samples, where $\tau$ is smoothing parameter. For the $k$-th neighbor $\mathbf{z}_n^k$ of $\mathbf{z}_n$, the opinions of $J$ experts are consolidated into $\dot{s}_n^k$ according to $\mathbf{J}_n^K$, where $\mathbf{J}_n^K[j]$ denotes the $j$-th index element of the $\mathbf{J}_n^K$. Note that we leverage the adaptive neighbor~\cite{zhang2024kfnn} $K^*\leq K$ for $\tilde{y}_n$ (i.e., Eq.~\ref{eq:label}), which adaptively selects a subset of neighbors according to peak confidence levels to prevent over-smoothing:
\begin{equation}
   \begin{aligned}
   K^*=\argmax_{k\in\{1,\cdots,K\}}\left(\text{max}(\frac{W_k\cdot\dot{\mathbf{s}}^k_n}{\sum_{k'\leq k}W_{k'}\cdot\dot{\mathbf{s}}^{k'}_n})-\text{second-max}(\frac{W_k\cdot\dot{\mathbf{s}}^k_n}{\sum_{k'\leq k}W_{k'}\cdot\dot{\mathbf{s}}^{k'}_n})\right).
   \end{aligned}
\end{equation}


\section{Experiment}
\subsection{Implementation Details}
\textbf{Dataset.}
We utilize two public and a real-world colon In-house\footnote{The approval was granted by the Ethics Review Board SMF-IRB-2024-007 and KH2024-059} dataset. The BReAst Carcinoma Subtyping (BRACS)~\cite{brancati2022bracs} comprises 547 WSIs, which are categorized into 7 classes representing various stages of breast carcinoma. We evaluate 4 types of molecular subtype prediction (Luminal A-B, HER2(+), and Triple Negative) using the Early Breast Cancer Core-Needle Biopsy (BCNB) dataset~\cite{xu2021predicting}, comprising 1,058 core-needle biopsy WSIs. The In-house training set consists of 698 WSI samples, resulting in a total of 4,188 annotations from a cohort of 6 experts. Each expert performed a soft assignment for every sample into 7 categories: Tubular Adenoma, Tubulovillous Adenoma, Traditional Serrated Adenoma, Hyperplastic Polyp, Sessile Serrated Lesion, Inflammatory Polyp, and Lymphoid Polyp, ensuring that the assigned probabilities sum to 1. A single consensus diagnosis was determined through experts' agreement for an additional 684 WSIs, which we use as the test set. While benchmarks have official splits, 20\% of the In-house data was randomly sampled for validation.
\\
\textbf{Synthesizing Experts.}
Following domain practices~\cite{chu2021learning,rodrigues2018deep,zhang2024iwbvt}, we generate subject expert training and validation labels for public datasets. We emulate diagnostic heterogeneity by partitioning the $\tilde{\textbf{z}}$ feature space into 10 clusters and assigning cluster-specific reliability scores to $J=6$ experts. Let $e_{cj}\sim\text{U(8,12)}$ denote the reliability score of expert $j$ for cluster $c$. A soft label is synthesized using a $s_{nj}\sim\text{Dirichlet}(e_{cj}{\times}y_n+0.1),\forall{n\in{c}}$. The Dirichlet parameters reflect local expert mastery, ensuring relative consensus varies across distinct regions.
\\
\textbf{Training Settings.} 
We utilize the Otsu algorithm~\cite{otsu1975threshold} to extract $\times{256}$ size patches from 1 Microns Per Pixel WSIs, which are transformed into instances via a pre-trained feature extractor~\cite{jaume2024multistain,lu2024avisionlanguage}. MILs are trained using the Adam optimizer~\cite{kingma2014adam} at a learning rate of $1e-4$ with cross-entropy loss, where we use fixed seeds across all experiments, utilizing an NVIDIA$^\circledR$ RTX A6000. For efficiency and fair comparison, the FAISS library~\cite{douze2025faiss} is leveraged for all clustering-based methods. We adopt TransMIL~\cite{shao2021transmil} and DTFD-MIL-AFS~\cite{zhang2022dtfd}, which have been extensively validated architecture by numerous works, as the backbone MIL. We employ $(K,\tau)=(10,0.1)$, where Sec.~\ref{section:further} provides a sensitivity analysis.
\\
\textbf{Comparison Methods.}
We introduce MV, SL, and Ensemble, which are initial approaches~\cite{sheng2008get}. For comparison, we employ milestones that have reported state-of-the-art performance. It includes the Expectation-Maximization method that formulates the cause of non-consensus as hidden states~\cite{whitehill2009whose}, a multi-branch architecture~\cite{rodrigues2018deep}, and a data correction~\cite{li2023neighborhood}. In addition, it integrates recent feature-reference methods based on clustering: IWBVT~\cite{zhang2024iwbvt} and KFNN~\cite{zhang2024kfnn}.

\begin{table}[t]
\centering
\caption{Qualitative results on various comparison methods and RaLMPH. $^\dagger$ and $^{\ddagger}$ indicates $p<0.05$ and $p<0.01$, respectively.
}
\label{tab:main}
\resizebox{\textwidth}{!}{%
\begin{tabular}{c|>{\hspace{0pt}}m{0.204\linewidth}|cc|cc|cc} 
\hline
\multirow{2}{*}{MIL} & \multirow{2}{*}{Method} & \multicolumn{2}{c|}{In-house} & \multicolumn{2}{c|}{BRACS \cite{brancati2022bracs}} & \multicolumn{2}{c}{BCNB \cite{xu2021predicting}} \\ 
\cline{3-8}
 & & Accuracy & AUC & Accuracy & AUC & Accuracy & AUC \\ 
\hline
\multirow{12}{*}{\rotatebox{90}{TransMIL \cite{shao2021transmil}}} 
 & MV & 0.616\tiny{0.025} & 0.920\tiny{0.019} & 0.466\tiny{0.029} & 0.788\tiny{0.015} & 0.413\tiny{0.023} & 0.674\tiny{0.007} \\
 & SL & \uline{0.619}\tiny{0.042} & 0.933\tiny{0.011} & 0.457\tiny{0.017} & 0.772\tiny{0.019} & 0.422\tiny{0.021} & 0.667\tiny{0.015} \\
 & Ensemble & 0.617\tiny{0.028} & \uline{0.935}\tiny{0.021} & 0.482\tiny{0.016} & 0.783\tiny{0.016} & \uline{0.441}\tiny{0.012} & 0.681\tiny{0.006} \\
 & GLAD \cite{whitehill2009whose} & 0.613\tiny{0.047} & 0.887\tiny{0.037} & \uline{0.485}\tiny{0.024} & 0.793\tiny{0.016} & 0.423\tiny{0.030} & 0.685\tiny{0.016} \\
 & DL-CL \cite{rodrigues2018deep} & 0.577\tiny{0.026} & 0.919\tiny{0.025} & 0.451\tiny{0.052} & 0.753\tiny{0.024} & 0.433\tiny{0.015} & 0.676\tiny{0.011} \\
 & NWVNC \cite{li2023neighborhood} & 0.533\tiny{0.016} & 0.861\tiny{0.014} & 0.473\tiny{0.035} & \uline{0.800}\tiny{0.016} & 0.440\tiny{0.020} & \uline{0.695}\tiny{0.020} \\
 & IWBVT \cite{zhang2024iwbvt} & 0.569\tiny{0.054} & 0.894\tiny{0.018} & 0.464\tiny{0.027} & 0.762\tiny{0.034} & 0.402\tiny{0.024} & 0.658\tiny{0.172} \\
 & KFNN \cite{zhang2024kfnn} & 0.574\tiny{0.014} & 0.900\tiny{0.001} & 0.457\tiny{0.036} & 0.783\tiny{0.020} & 0.393\tiny{0.029} & 0.679\tiny{0.016} \\
 & \cellcolor[rgb]{0.937,0.937,0.937}RaLMPH & \cellcolor[rgb]{0.937,0.937,0.937}\textbf{0.627}\tiny{0.012} & \cellcolor[rgb]{0.937,0.937,0.937}\textbf{0.941}\tiny{0.008} & \cellcolor[rgb]{0.937,0.937,0.937}\textbf{0.494}\tiny{0.016} & \cellcolor[rgb]{0.937,0.937,0.937}\textbf{0.807}\tiny{0.019} & \cellcolor[rgb]{0.937,0.937,0.937}\textbf{0.460}\tiny{0.022} & \cellcolor[rgb]{0.937,0.937,0.937}{\textbf{0.712}\tiny{0.012}}$^{\ddagger}$ \\
 & \cellcolor[rgb]{0.941,0.867,0.659}\hfill \textit{w.o.} $\Phi$ & \cellcolor[rgb]{0.941,0.867,0.659}0.605\tiny{0.045} & \cellcolor[rgb]{0.941,0.867,0.659}0.934\tiny{0.011} & \cellcolor[rgb]{0.941,0.867,0.659}0.458\tiny{0.029} & \cellcolor[rgb]{0.941,0.867,0.659}0.774\tiny{0.020} & \cellcolor[rgb]{0.941,0.867,0.659}0.432\tiny{0.018} & \cellcolor[rgb]{0.941,0.867,0.659}0.700\tiny{0.004} \\
 & \cellcolor[rgb]{0.941,0.867,0.659}\hfill \textit{w.o.} $\mathbf{J}_n^k$ & \cellcolor[rgb]{0.941,0.867,0.659}0.610\tiny{0.039} & \cellcolor[rgb]{0.941,0.867,0.659}0.928\tiny{0.013} & \cellcolor[rgb]{0.941,0.867,0.659}0.464\tiny{0.049} & \cellcolor[rgb]{0.941,0.867,0.659}0.780\tiny{0.025} & \cellcolor[rgb]{0.941,0.867,0.659}0.438\tiny{0.016} & \cellcolor[rgb]{0.941,0.867,0.659}0.701\tiny{0.008} \\
 & \cellcolor[rgb]{0.941,0.867,0.659}\hfill \textit{w.o.} Adapt. $K^*$ & \cellcolor[rgb]{0.941,0.867,0.659}0.604\tiny{0.030} & \cellcolor[rgb]{0.941,0.867,0.659}0.926\tiny{0.009} & \cellcolor[rgb]{0.941,0.867,0.659}0.478\tiny{0.033} & \cellcolor[rgb]{0.941,0.867,0.659}0.762\tiny{0.034} & \cellcolor[rgb]{0.941,0.867,0.659}0.454\tiny{0.024} & \cellcolor[rgb]{0.941,0.867,0.659}0.705\tiny{0.024} \\ 
\hline
\multirow{12}{*}{\rotatebox{90}{DTFD-MIL-AFS \cite{zhang2022dtfd}}} 
 & MV & 0.665\tiny{0.041} & 0.904\tiny{0.023} & 0.512\tiny{0.023} & \uline{0.844}\tiny{0.005} & \uline{0.513}\tiny{0.025} & 0.759\tiny{0.009} \\
 & SL & 0.697\tiny{0.029} & \uline{0.913}\tiny{0.005} & 0.478\tiny{0.029} & 0.826\tiny{0.007} & 0.511\tiny{0.017} & 0.761\tiny{0.008} \\
 & Ensemble & 0.680\tiny{0.026} & \textbf{0.923}\tiny{0.006} & \uline{0.521}\tiny{0.011} & 0.838\tiny{0.010} & 0.511\tiny{0.019} & 0.764\tiny{0.007} \\
 & GLAD \cite{whitehill2009whose} & \uline{0.708}\tiny{0.020} & 0.890\tiny{0.026} & 0.515\tiny{0.019} & \textbf{0.845}\tiny{0.005} & 0.503\tiny{0.017} & 0.759\tiny{0.014} \\
 & DL-CL \cite{rodrigues2018deep} & 0.703\tiny{0.038} & 0.909\tiny{0.017} & 0.496\tiny{0.031} & 0.833\tiny{0.005} & 0.503\tiny{0.024} & \uline{0.770}\tiny{0.003} \\
 & NWVNC \cite{li2023neighborhood} & 0.603\tiny{0.027} & 0.901\tiny{0.015} & 0.510\tiny{0.029} & 0.819\tiny{0.011} & 0.497\tiny{0.014} & 0.750\tiny{0.009} \\
 & IWBVT \cite{zhang2024iwbvt} & 0.662\tiny{0.032} & 0.908\tiny{0.015} & 0.478\tiny{0.011} & 0.826\tiny{0.007} & 0.499\tiny{0.014} & 0.758\tiny{0.009} \\
 & KFNN \cite{zhang2024kfnn} & 0.644\tiny{0.016} & 0.896\tiny{0.016} & 0.459\tiny{0.012} & 0.837\tiny{0.006} & 0.471\tiny{0.012} & 0.750\tiny{0.003} \\
 & \cellcolor[rgb]{0.937,0.937,0.937}RaLMPH & \cellcolor[rgb]{0.937,0.937,0.937}\textbf{0.718}\tiny{0.021} & \cellcolor[rgb]{0.937,0.937,0.937}\uline{0.913}\tiny{0.014} & \cellcolor[rgb]{0.937,0.937,0.937}{\textbf{0.534}\tiny{0.014}}$^\dagger$ & \cellcolor[rgb]{0.937,0.937,0.937}\textbf{0.845}\tiny{0.006} & \cellcolor[rgb]{0.937,0.937,0.937}\textbf{0.533}\tiny{0.037} & \cellcolor[rgb]{0.937,0.937,0.937}{\textbf{0.781}\tiny{0.009}}$^\ddagger$ \\
 & \cellcolor[rgb]{0.941,0.867,0.659}\hfill \textit{w.o.} $\Phi$ & \cellcolor[rgb]{0.941,0.867,0.659}0.700\tiny{0.031} & \cellcolor[rgb]{0.941,0.867,0.659}0.907\tiny{0.018} & \cellcolor[rgb]{0.941,0.867,0.659}0.517\tiny{0.018} & \cellcolor[rgb]{0.941,0.867,0.659}0.837\tiny{0.011} & \cellcolor[rgb]{0.941,0.867,0.659}0.507\tiny{0.014} & \cellcolor[rgb]{0.941,0.867,0.659}0.768\tiny{0.009} \\
 & \cellcolor[rgb]{0.941,0.867,0.659}\hfill \textit{w.o.} $\mathbf{J}_n^k$ & \cellcolor[rgb]{0.941,0.867,0.659}0.678\tiny{0.013} & \cellcolor[rgb]{0.941,0.867,0.659}0.905\tiny{0.021} & \cellcolor[rgb]{0.941,0.867,0.659}0.521\tiny{0.020} & \cellcolor[rgb]{0.941,0.867,0.659}0.836\tiny{0.012} & \cellcolor[rgb]{0.941,0.867,0.659}0.499\tiny{0.020} & \cellcolor[rgb]{0.941,0.867,0.659}0.768\tiny{0.004} \\
 & \cellcolor[rgb]{0.941,0.867,0.659}\hfill \textit{w.o.} Adapt. $K^*$ & \cellcolor[rgb]{0.941,0.867,0.659}0.705\tiny{0.020} & \cellcolor[rgb]{0.941,0.867,0.659}0.909\tiny{0.017} & \cellcolor[rgb]{0.941,0.867,0.659}0.521\tiny{0.023} & \cellcolor[rgb]{0.941,0.867,0.659}0.838\tiny{0.011} & \cellcolor[rgb]{0.941,0.867,0.659}0.520\tiny{0.011} & \cellcolor[rgb]{0.941,0.867,0.659}0.778\tiny{0.006} \\
\hline
\end{tabular}
}
\end{table}

\subsection{Quantitative Performance}
\textbf{Comparison Results.} Table \ref{tab:main} summarizes the performance comparison on three datasets. MV and SL serve as competitive baselines, while Ensembling yields higher overall AUC at the expense of increased computational time, which is analyzed in Sec.~\ref{section:further}. GLAD, which leverages hidden states, demonstrates superior performance over structural~\cite{rodrigues2018deep} and data correction~\cite{li2023neighborhood} approaches. The performance decline in IWBVT, which data features are not considered, suggests the necessity of jointly modeling label information and feature representations. In contrast, the underperformance of KFNN suggests that single-instance assumptions are ill-suited for multi-instance WSI analysis. Our proposed RaLMPH consistently outperforms all competitive methods across two MIL backbones. Its robust performance on both simulated benchmarks and clinical datasets underscores its practical utility in real-world pathology.
\\
\textbf{Ablation Study.} In Tab.~\ref{tab:main}, the values in the background \crule[myGold]{0.25cm}{0.25cm} show the ablation results on the RaLMPH. Ablation $\Phi$ (i.e., L$_2$ distance) results in a performance decline, highlighting the validation of the joint instance bag variance and the entropy of the label. Excluding $\mathbf{J}^k_n$ also degrades the results, especially for DTFD-MIL, while adaptive $K^*$ ensures a more robust stability. These findings underscore that each component is integral to achieving optimal results.

\subsection{Further Analysis}\label{section:further}
\begin{figure}[t!]
\centering
\includegraphics[width=1\linewidth]{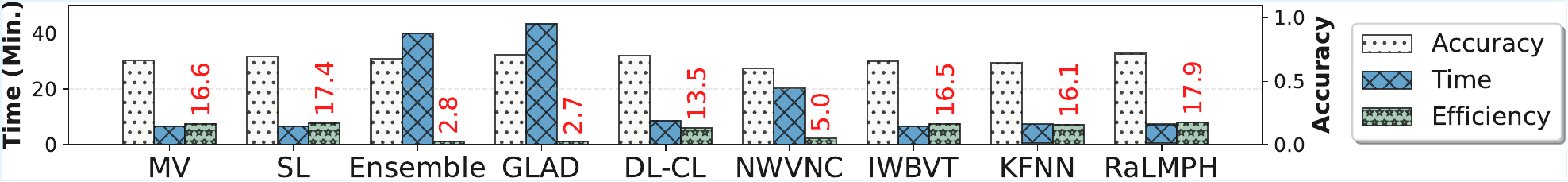}
\caption{
Efficiency of various comparisons and RaLMPH. The efficiency is calculated as $100\times\text{(Accuracy/}(60\times\text{Time}))$.
}
\label{fig:effi}
\end{figure}

\begin{figure}[t!]
\centering
\includegraphics[width=1\linewidth]{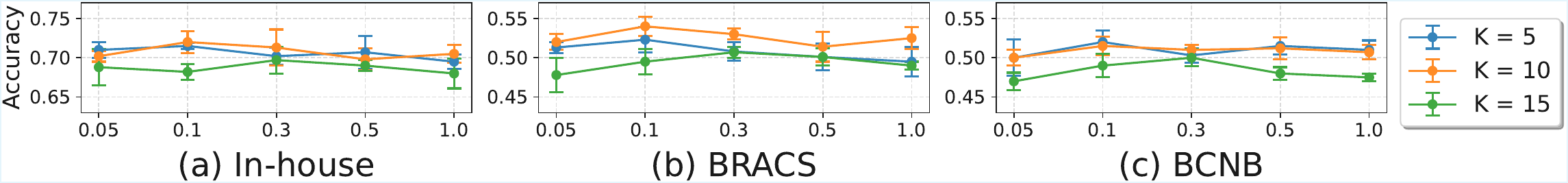}
\caption{
Sensitivity analysis of $K$ and $\tau$, with $\tau$ plotted on the $x$-axis.
}
\label{fig:sensi}
\end{figure}

\textbf{Qualitative Observation on $\Phi$.}
\begin{figure}[t!]
\centering
\includegraphics[width=1\linewidth]{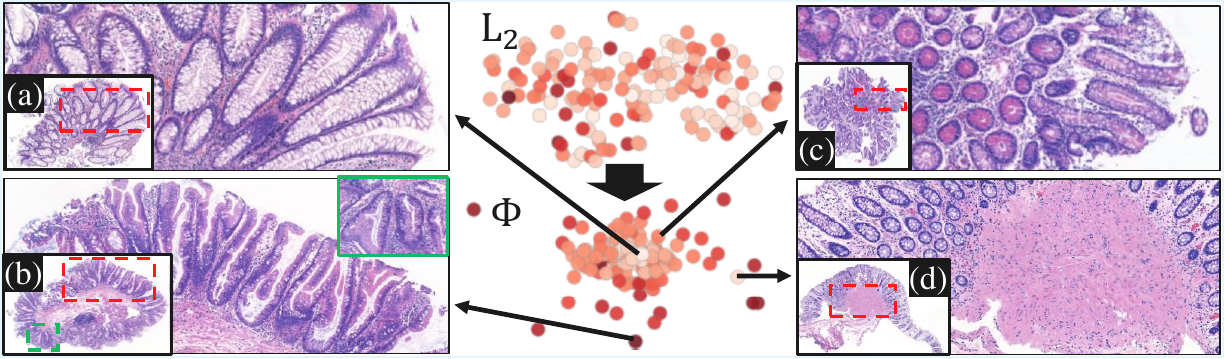}
\caption{
TA sample clusters in L$_2$ and $\Phi$ distance alongside WSI examples.
}
\label{fig:phi}
\end{figure}
Fig.~\ref{fig:phi} visualizes the Tubular Adenoma (TA), which was 
temporarily determined by MV, sample clusters in the L$_2$ and $\Phi$ distance spaces~\cite{maaten2008visualizing}, with a color intensity representing high $\mathcal{H}(\bar{\mathbf{s}}_n)$, i.e., low label consensus. L$_2$ distance leaves expert opinions scattered as it ignores $\bar{\mathbf{s}}_n$ and bag-level dynamics, whereas $\Phi$ aligns it by jointly modeling $\mathcal{H}(\bar{\mathbf{s}}_n)$ and $\bm{\sigma}^2_n$. Case (a) presents a typical TA with elongated glands, exhibiting low $\mathcal{H}(\bar{\mathbf{s}}_n)$ and $\bm{\sigma}^2_n$. Conversely, (b) displays high $\mathcal{H}(\bar{\mathbf{s}}_n)$ and $\bm{\sigma}^2_n$ due to the overlap features of the superficial serration and tubular glands. Case (c) shows separated opinions among experts, caused by dense inflammation and partial tubulation. In particular, the intentionally included Leiomyoma (d) achieved low $\mathcal{H}(\bar{\mathbf{s}}_n)$ as recognized an outlier by experts, yet it is located in the periphery due to its high $\bm{\sigma}^2_n$.
\\
\textbf{Robustness on the Number of Experts.}
\begin{figure}[t!]
\centering
\includegraphics[width=1\linewidth]{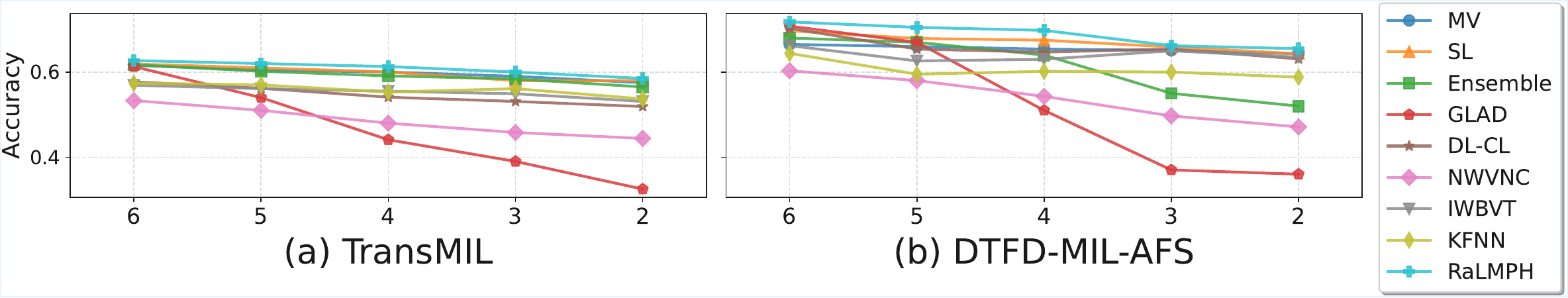}
\caption{
Performance variation with respect to $J'$, which is plotted on the $x$-axis.
}
\label{fig:J}
\end{figure}
We evaluate robustness against a reduced number of experts (i.e., $J'<J$) on the In-house dataset. Performance across all $J'$ is measured by averaging multiple runs with $J-J'$ randomly omitted annotators (Fig.~\ref{fig:J}). While Ensemble shows performance drops at lower $J'$, MV and SL baselines remain robust. As $J'$ decreases, merging strategies such as IWBVT tend to converge toward the baselines' performance. GLAD collapses as $J'$ decreases, indicating the difficulty of parameter estimation with limited labels. NWVNC also exhibits significant degradation, limiting its practical utility. RaLMPH maintains stability without substantial deductions, providing empirical evidence of its readiness for real-world clinical deployment.
\\
\textbf{Time Efficiency.}
Fig.~\ref{fig:effi} shows the time efficiency of various comparisons and RaLMPH. MV, SL, Ensemble, and GLAD have short preprocessing time, as they either bypass label refinement or require only $\mathcal{O}(1)$ complexity. However, the Ensemble necessitates a linear increase in time proportional to $J$ annotators. While GLAD and NWVNC require iterative refinement steps, DL-CL achieves the benefits of Ensembling in significantly less time through a parallelized architecture. Methods involving early-stage refinement, such as IWBVT, KFNN, and RaLMPH, introduce no additional overhead during training; among these, RaLMPH demonstrates superior efficiency in practical applications.
\\
\textbf{Sensitivity Analysis.}
We provide sensitivity analyses using DTFD-MIL with respect to $K$ and $\tau$ in Fig.~\ref{fig:sensi}. The performance degradation at $K=15$ underscores the negative impact of over-smoothing. In contrast, $K=5$ and $K=10$ performed similarly, with $K=10$ being marginally better. The adaptive $K^*$ strategy ensures reliable neighbor selection provided the initial pool is reasonable. We found $\tau=0.1$ to be the most effective, while overly large $\tau$ values cause abrupt performance drops. These findings establish a stable parameter baseline that can be applied to various datasets.

\section{Conclusion}
In this paper, we address the critical gap between conventional MIL assumptions and the inherent inter-expert variability in the clinical practice WSI dataset. We introduce RaLMPH, a novel label reconciliation framework that moves beyond global reliability metrics by focusing on localized diagnostic authority. By incorporating a reliability field that jointly models WSI feature variance and expert entropy, RaLMPH adaptively identifies trustworthy neighborhoods and prioritizes the most reliable opinions for each sample. Our experiments on clinical datasets with six pathologists and simulated benchmarks demonstrate that RaLMPH consistently outperforms state-of-the-art methods across different MIL backbones. These results underscore the importance of accounting for local expertise in multi-annotator environments, offering a more robust and clinically relevant approach to automated pathology analysis.
\\
\noindent\textbf{Limitations and Future Work.}
Our current scope assumes the involvement of designated experts and is not readily extendable to crowdsourced annotations. Future work will investigate more generalized scenarios involving irregular clinical participation, while integrating advanced MIL frameworks.
\\
\noindent\textbf{Acknowledgment.} 
This work was supported by the National Research Foundation of Korea (NRF) grant funded by the Korean government (MSIT) under grant number RS-2022-NR068758. We are also deeply grateful for the generous support provided by the Seegene Medical Foundation in South Korea.
\\
\noindent\textbf{Disclosure of Interests.} The authors have no competing interests to declare that are relevant to the content of this article.
\bibliographystyle{splncs04}
\bibliography{0.main}

\end{document}